# Interpretable Convolutional Neural Networks for Subject-Independent Motor Imagery Classification

Ji-Seon Bang
*Dept. Brain and Cognitive Engineering*
Korea University
Seoul, Republic of Korea
js bang@korea.ac.kr

Seong-Whan Lee
*Dept. Artificial Intelligence*
Korea University
Seoul, Republic of Korea
sw.lee@korea.ac.kr

*Abstract*—Deep learning frameworks have become increasingly popular in brain computer interface (BCI) study thanks to their outstanding performance. However, in terms of the classification model alone, they are treated as black box as they do not provide any information on what led them to reach a particular decision. In other words, we cannot convince whether the neuro-physiological factor or simply noise is the factor of high performance. Because of this disadvantage, it is difficult to ensure adequate reliability compared to their high performance. In this study, we propose an explainable deep learning model for BCI. Specifically, we aim to classify EEG signal which is obtained from the motor-imagery (MI) task. In addition, we adopted layer-wise relevance propagation (LRP) to the model to interpret the reason that the model derived certain classification output. We visualized the heatmap which indicates the output of the LRP in form of topography to certify neuro-physiological factors. Furthermore, we classified EEG with the subject-independent manner to learn robust and generalized EEG features by avoiding subject dependency. The methodology also provides the advantage of avoiding the expense of building training data for each subject. With our proposed model, we obtained generalized heatmap patterns for all subjects. As a result, we can conclude that our proposed model provides neuro-physiologically reliable interpretation.

*Keyworkds*—brain–computer interface, electroencephalography, convolutional neural network, motor imagery, layer-wise relevance propagation, explainable artificial intelligence

## I. INTRODUCTION

Brain computer interface (BCI) refers to a system capable of controlling an external device by decoding a user's intention or cognitive states from their brain signal [1]–[4]. Electroencephalography (EEG) which is a non-invasive brain signal recording method is widely used for BCI communication approach. There are several paradigms such as steady-state visually evoked potential (SSVEP) [5]–[7], event-related potential (ERP) [8], [9], and rapid serial visual presentation (RSVP) [10], [11]. Especially, motor-imagery (MI) [12]–[14] task is one of the main BCI paradigms that enable the aim of BCI. The MI signal is derived when the user mentally simulates or imagines certain movement by a given cue.

To decode EEG signal, the deep learning approach is widely used in recent years as they show enhanced performance compared to previous machine learning methods. Sturm *et al.* [15], Schirrmeister *et al.* [16], and Sakhavi *et al.* [17], and Bang *et al.* [18] classified EEG signal from MI task with deep learning-based model. The first was a deep neural network (DNN) based approach, and the other three were a convolutional neural network (CNN) based approaches. Among them, Sturm *et al.* [15] and Bang *et al.* [18] proposed classification models as well as interpretation methods. Without the interpretation method, we cannot assure whether high performance is due to neuro-physiological factors or simply to noise. By providing the interpretation method, the reliability of the model can be further improved.

In a study by Sturm *et al.* [15], the interpretation method was first applied in BCI study field using the layer-wise relevance propagation (LRP) method. Bang *et al.* [18] also used the LRP method to explain the decisions predicted by the proposed classifier. LRP algorithm was developed to explain the decision of classifier specific to a given input data. This is achieved by the concept of relevance scores. LRP attributes relevance scores to important components of the input data by using the topology of the classifier itself. This is possible because LRP operates with the back-propagation method.

Meanwhile, a process called calibration is required to use the BCI system. calibration refers to a process that collects custom training data for users who will use the BCI system. Since EEG has large inter-subject and intra-subject variability, it is hard to use other subject's signals as training data because of the performance reduction. To overcome this limitation, the concept of subject-independent was proposed. This method minimizes performance degradation or rather improves performance by collecting data from multiple subjects and then learning it to extract generalized feature representation. With the subject-independent method, one can use BCI system in real-time without having to first train the classifier using the user's subject-specific data [19]–[21].

In this study, the MI classification model was proposed and the results are visually interpreted in the form of topography with LRP. Also, by adopting the subject-independent method

This work was supported in part by the Institute for Information & Communications Technology Promotion (IITP) grant, funded by the Korea government (MSIT) (No. 2015-0-00185, Development of Intelligent Pattern Recognition Softwares for Ambulatory Brain Computer Interface, No. 2017-0-00451, Development of BCI based Brain and Cognitive Computing Technology for Recognizing User's Intentions using Deep Learning, No. 2019-0-00079, Artificial Intelligence Graduate School Program, Korea University) in the process of evaluation, we obtained generalized topography patterns from the interpreted results. Specifically, we filtered raw EEG signals with six different frequency bands and obtained maximum and minimum value of amplitude from each segmented signals. We projected three-dimensional channel location into two-dimensional map and assigned the obtained maximum and minimum value to each channel location. Through the process, we obtained (6 × 7) sized 12 maps. We classified them with four-layered CNN classifier and obtained relevance scores with the LRP model. We visualized the relevance scores, which represents the highly contributing channels, by projecting each channel value into the topography. The whole process has proceeded with the subject-independent manner. The proposed method achieved 68.13% accuracy and was able to explain the reason for the classification. With the proposed method, it is possible to get high performance and enhance model reliability.

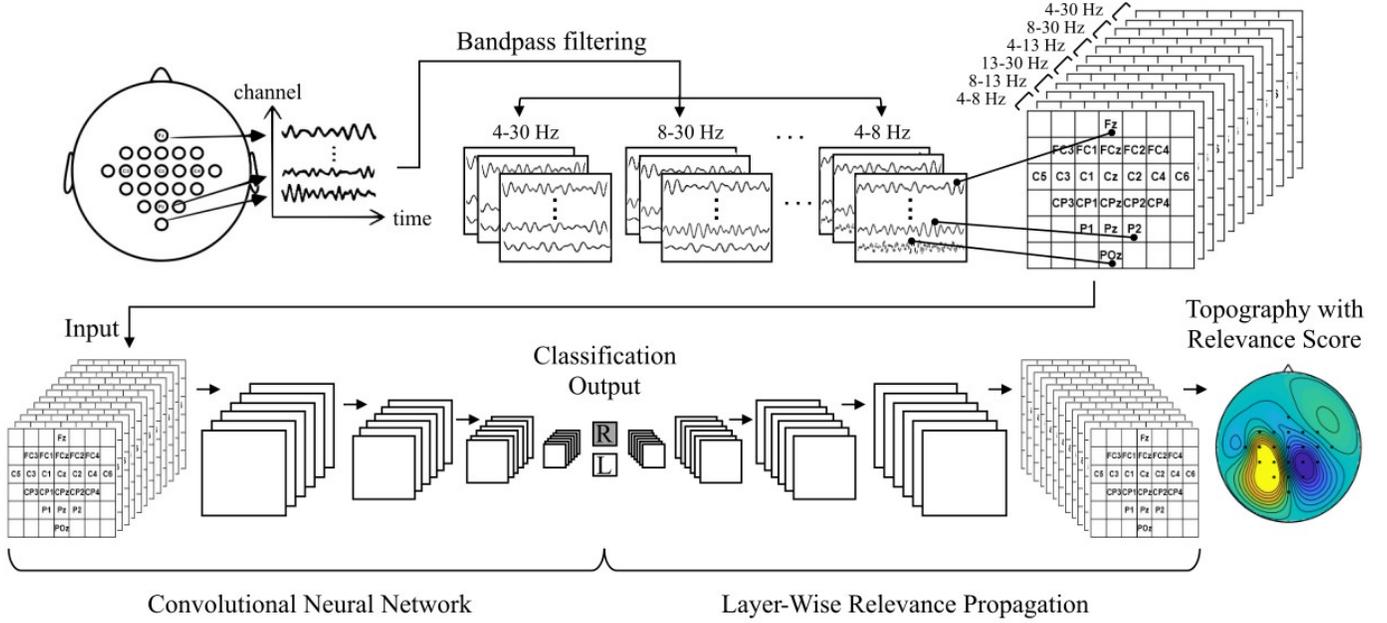

Fig. 1. The framework of the proposed method to classify and visually interpret motor imagery EEG signal.

## II. METHOD

In this section, we will introduce the overall framework of the proposed method. Specifically, we will first focus on how to generate an input map and then will introduce our classification model and interpretation model. Fig. 1 presents the overview of the proposed framework. Further detailed information will be explained below.

### A. Generating Feature Map

For the first process of generating feature map, we band-pass filtered raw signal with 6 frequency bands, which are $[4, 8], [8, 13], [13, 30], [4, 13], [8, 30], [4, 30]$. Band-pass filtering is a method to specify the signal of interest. The selected bands are $\theta$, $\alpha$, and $\beta$, $\theta + \alpha$, $\alpha + \beta$, and $\theta + \alpha + \beta$ bands respectively. They are selected because $\theta$, $\alpha$, and $\beta$ bands are commonly used bands for MI analysis and classification. Then we segmented each filtered signal with two-second interval; between 0.5 s and 2.5 s after the onset of the cue. Six segmented signals were normalized with a function named local average reference individually.

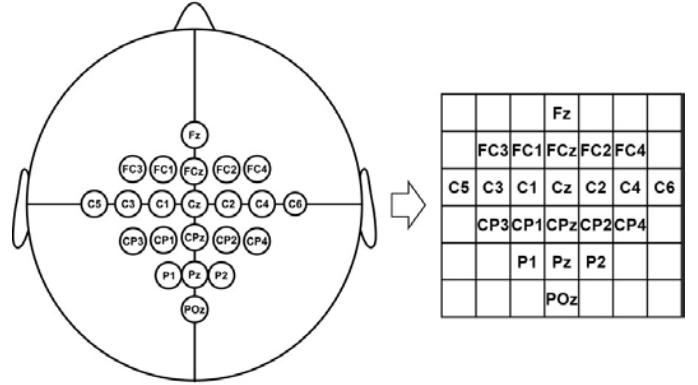

Fig. 2. The formation of input map.

Meanwhile, we mapped the three-dimensional location of channels into the two-dimensional map. Fig. 2 shows the mapped locations of 28 channels which are the entire channels of the data we used. We obtained the maximum and minimum values for each two-second length trial. Then we mapped the obtained maximum and minimum values into two-dimensional map. This process was done equally for six filtered signals. As a result, 12 maps were obtained with the entire process. These are the input feature for the CNN classifier, whose size is $6 \times 7 \times 12$. Here, two-dimensional ($6 \times 7$) sized maps were used as an actual input, and the third dimension was treated as a feature map like RGB in image classification.

### B. Convolutional Neural Network

The CNN network consists of 4 convolutional layers with the rectified linear unit (RELU) [22] layer and 1 fully-connected layer. The filter sizes are $(2 \times 2)$, $(2 \times 2)$, $(2 \times 2)$, and $(3 \times 4)$ respectively. As we used 'VALID' padding for all four layers, the size of feature maps is shrunken into $1 \times 1$ at the end of the last layer. The size of feature maps is set to 32 for all four layers. As a result, the $((6 \times 7) \times 12)$ sized input changes into $((5 \times 6) \times 32)$, $((4 \times 5) \times 32)$, $((3 \times 4) \times 32)$, and finally $((1 \times 1) \times 32)$. The last fully-connected layer changes into $((1 \times 1) \times 2)$, as the model is a binary classifier. We employed Adam-optimizer [23] to optimize the cost. The learning rate was set to 0.0001. We obtained the decoding accuracy when the number of iteration reached 300.

### C. Layer-Wise Relevance Propagation

To visually interpret the analysis of the classification result, we applied an LRP method. The idea of the LRP is to find the positive and negative contribution of each input pixel made by the classifier. To determine the relevance scores, we assume that the classification output is the prediction score. Then, the LRP decomposes the classification output into sum of feature relevance scores with back-propagation. As LRP obtains relevance scores by back-propagating, the shape of the model is opposite to the CNN structure. The $((1 \times 1) \times 2)$ sized input of LRP, which is also the output of CNN changes into $((1 \times 1) \times 32)$, $((3 \times 4) \times 32)$, $((4 \times 5) \times 32)$, and $((5 \times 6) \times 32)$, and finally $((6 \times 7) \times 12)$. The final relevance scores are derived in the form of input size which is $6 \times 7 \times 12$. We averaged the third dimension and mapped the two-dimensional output into the topography.

## III. DATA DESCRIPTION AND EVALUATION

### A. Evaluation Data

BCI Competition IV_2a data [24] which is commonly used in MI studies was adopted for the verification. The data was measured when nine subjects, namely (A01–A09) performed MI tasks with given cue which are displayed on the monitor. Either right or left arrow is displayed as a cue, and the subject imagines hand movement of the corresponding direction the cue points to. It is possible to classify EEG signals as brain signals vary depending on which hand movement the subject imagines. Although the original data consist of 25 channels including 3 EOG channels, we only used 22 EEG channels in this study. Also, the data includes four classes but only left and right-hand motor imagery classes are selected for this study. The performance of the proposed method was evaluated by the accuracy with the subject-independent method.

### B. Evaluation Method

We used the subject-independent method for evaluation. By adopting the method, calibration time would greatly reduced, and generalized features would be extracted with the proposed classification and interpretation model. We will introduce how we assigned training data and test data to adopt the subject-independent method.

In BCI competition IV_2a data which we used for the evaluation, training set, and test set are already separated for every nine subjects. Here, we only used the test set of the subject who we are going to evaluate as the test data. As for the training data, the entire training set and test set for the remaining eight subjects were used except for the subject to be evaluated. Fig. 3 shows the example of when subject 3 is evaluated. In here, the first and second row indicate a training set and a test set in the IV_2a data, respectively. The boxes colored with dark grey and light grey present the data which we used for training and test for verifying proposed method, respectively.

We compared the classification accuracy of our model with the baseline method. In this study, common spatial pattern

Fig. 3. Allocation of training data and test data for a subject-independent method.

| Data for training: ■ | Data for test: ▢ |

| IV_2a data | Subjects | | | | | | | | |
|---|---|---|---|---|---|---|---|---|---|
| Training set | A01T | A02T | A03T | A04T | A05T | A06T | A07T | A08T | A09T |
| Test set | A01E | A02E | A03E | A04E | A05E | A06E | A07E | A08E | A09E |

TABLE I
COMPARISON OF THE PROPOSED METHOD AND BASELINE METHOD.

| Subjects | CSP [25] | Proposed |
|---|---|---|
| A01 | 61.11 | 88.19 |
| A02 | 53.47 | 61.11 |
| A03 | 89.58 | 79.17 |
| A04 | 52.08 | 60.42 |
| A05 | 51.39 | 55.56 |
| A06 | 47.92 | 64.58 |
| A07 | 52.78 | 59.72 |
| A08 | 87.50 | 86.11 |
| A09 | 60.42 | 58.33 |
| mean | 61.81 | 68.13 |

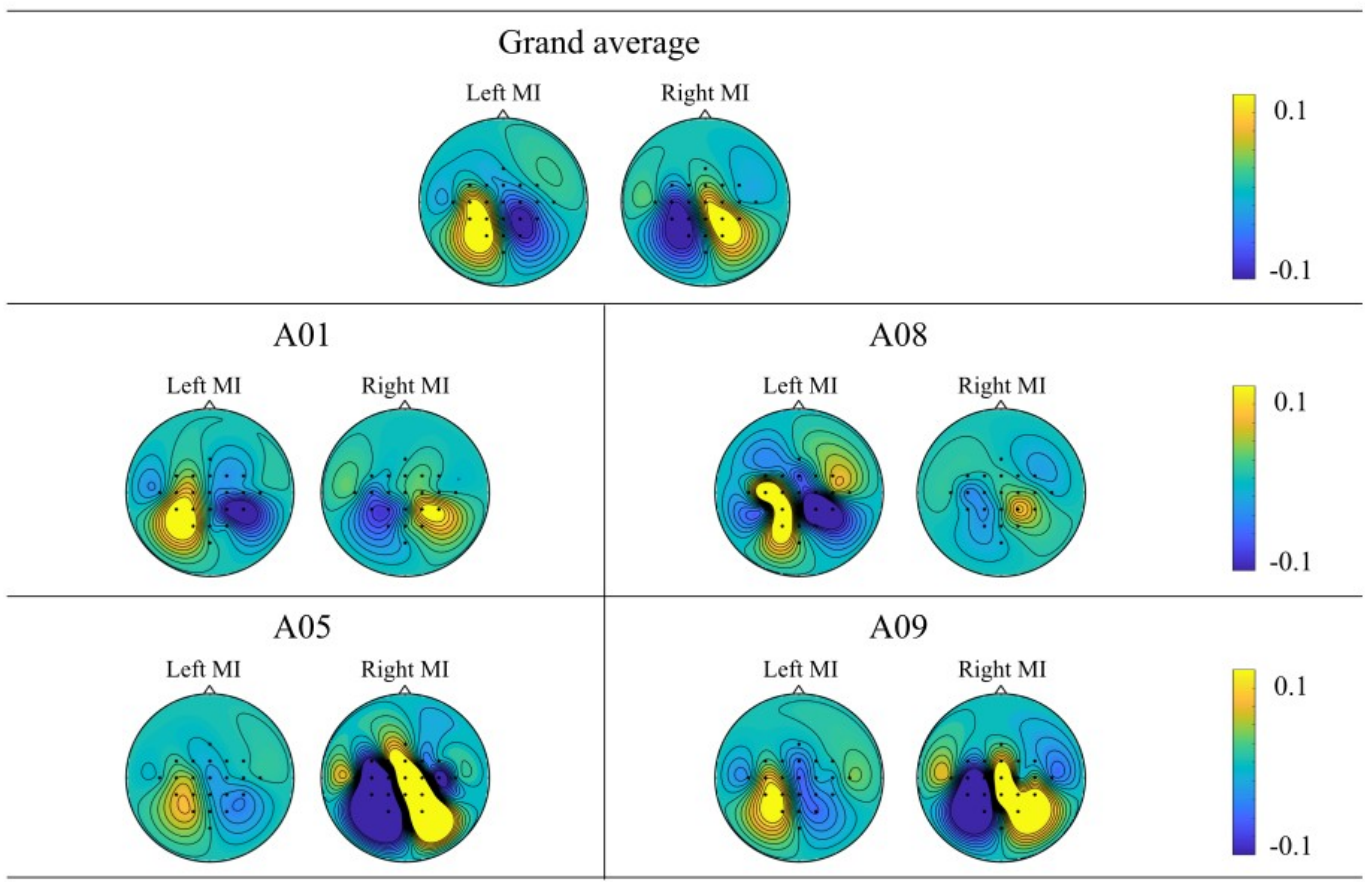

Fig. 4. Topography of the relevance scores. We plotted the topography of grand average, the two subjects A01 and A08 who achieved the highest decoding accuracy, and the two subjects A05 and A09 who achieved the lowest decoding accuracy.

(CSP) [25] with linear discriminant analysis (LDA) [26] was adopted for baseline method. The signal was band-pass filtered between 8 and 12 Hz ($\mu$ band). For the pre-processing, the signal was segmented between 0.5 and 2.5 s after the onset of the cue as same as the proposed method. The evaluation was also conducted in a subject-independent method. Training data and test data were selected in the same way as the proposed method.

## IV. RESULT

TableI represents the performance that was obtained with the BCI Competition IV 2a data. Results of both proposed method and baseline method are obtained with subject-independent method. The mean accuracy of the baseline method was 61.81% and the accuracy of the proposed model was 68.13%. The proposed method outperformed the baselines method.

After the decoding accuracy was obtained, the outputs were back-propagated to yield relevance scores with LRP. Fig. 4 present the topography of the relevance scores that indicates which channels have the highest and lowest contribution. Including grand average, subjects A01 and A08, who obtained the highest (88.19%) and second highest (86.11%) decoding accuracy, and subjects A05 and A09 who obtained the lowest (55.56%) and second lowest (58.33%) decoding accuracy are presented. The scores were obtained by averaging the accurately identified trials in each class. Also, the feature maps in third dimension are averaged so that ($6 \times 7$) sized heatmaps are finally obtained. We assigned the values corresponding to each channel to the topography. The topographies of relevance scores were plotted in the same range, -0.1 to 0.1, across subjects.

## V. DISCUSSION

In Fig. 4, the left and right hemispheres are distinguishable for all plotted subjects. The left hemisphere has a high relevance scores when imagining the left hand movement and vice versa. This result represent that the left hemisphere contributes more than another side of the brain in the left imaging MI task. In general, C3 and C4 channels are known to have the highest contribution in MI task, as they are located in the motor cortex areas. From our results, it can be confirmed that it is consistent with the existing neuro-physiological knowledge. In particular, similar patterns appear not only in subjects who achieved high performance but also in subjects who achieved low performance. As a result, we can conclude that our interpretation model produces neuro-physiologically plausible explanations of how the proposed model reaches a certain decision.

In this study, a novel framework that can classify MI as well as interpret the output of the classifier was proposed. The proposed method was able to explain the reason for the classification. LRP produces neuro-physiologically reasonable explanations of how the classifier reaches decisions. Also, Table I shows that the proposed method reached higher decoding accuracy compared to the baseline method when compared with BCI Competition IV 2a data. The proposed method could be applied to other research areas such as BCI assistive application [27], [28] or magnetic resonance imaging (MRI) study [29], [30].